\documentclass[runningheads]{llncs}
\usepackage[T1]{fontenc}
\usepackage{graphicx,verbatim}
\usepackage{booktabs} 
\usepackage{threeparttable}
\usepackage{multirow}
\usepackage{graphicx}
\usepackage{subcaption}
\usepackage{amsmath}

\usepackage{amssymb}
\usepackage{tabularx}
\usepackage{makecell}
\usepackage{array}
\begin{document}
\title{Fusion is not one-size-fits-all: Cross-Modal Representation Alignment for Time-to-Event Modeling}
%

\author{Zhemin Zhang\inst{1} \and
Weijie Chen\inst{2} \and
David Le\inst{2} \and
Amara Tariq\inst{2} \and
Alex Wallace\inst{2} \and
Matthew Stib\inst{2} \and
Juan Maria Farina\inst{2} \and
Chadi Ayoub\inst{2} \and
Reza Arsanjani\inst{2} \and 
Imon Banerjee\inst{1,2}  
\authorrunning{Zhang et al.}
\institute{Arizona State University \\
    \and
    Mayo Clinic}
  }
\maketitle              
\begin{abstract}
Accurate time-to-event (TTE) prediction from multimodal clinical data remains challenging due to modality imbalance and distribution shift. We introduce a foundation model–driven framework for cross-modal representation alignment between CT imaging and longitudinal EHR data, designed to generalize across tasks and institutions. CT and EHR modalities are encoded independently using domain-specific foundation models and aligned in a shared latent space through four principled fusion strategies: late fusion, contrastive alignment, cross-attention, and co-attention. We evaluate two clinically distinct TTE tasks—pulmonary embolism (PE) mortality and cardiovascular disease (CVD) outcomes—on large-scale multi-institutional cohorts (PE: N=3,099 train; 1,098 internal; 435 external; CVD: N=2,951 train; 837 internal; 682 external). Fusion consistently improves concordance index by 1.5–5.4\% over unimodal baselines when modalities contribute comparably. 
Overall, contrastive multimodal fusion—particularly with CLMBR representations—provided the most consistent and statistically robust improvements, especially for PE mortality prediction. For MACE, cross-attention (one-hot) achieved the highest internal performance and image-guided co-attention achieved the best external performance. We therefore introduce a generalizable foundation model–based cross-modal alignment framework and provide the first systematic analysis of fusion behavior under modality imbalance in TTE prediction. Our results establish task-aware multimodal alignment as a necessary design principle for robust generalization and scalable clinical deployment.

\keywords{Time-to-event prediction   \and Multimodal \and Foundation model.}

\end{abstract}
%
\section{Introduction}

Prognosis in healthcare requires estimating both the likelihood and timing of adverse outcomes. For high-risk conditions such as pulmonary embolism (PE) and cardiovascular disease (CVD), patient trajectories are heterogeneous, and accurate temporal risk stratification is essential for guiding monitoring, intervention, and resource allocation. Time-to-event (TTE) modeling provides a natural framework for estimating individualized risk over time, unlike binary classification approaches that predict only event occurrence~\cite{wiegrebe2024_survival_review,steinberg2024_motor_iclr}. Traditional risk scores rely on static tabular variables and often omit high-dimensional imaging data and longitudinal context~\cite{stone2022_cps_accuracy,khan2024_prevent_circulation}. In PE, scores such as Pulmonary Embolism Severity Index (sPESI) show variable calibration, while emerging evidence suggests that combining CT Pulmonary Angiography (CTPA) with clinical data can improve outcome prediction~\cite{zhang2024_spesi_validation_ejim,inspect2023_dataset}. Combining traditional risk score with CT derived biomarkers can improve prognostic performance~\cite{huang2020fusion}.

Multimodal deep learning offers a principled approach to integrate complementary prognostic information: imaging captures structural and spatial markers of disease severity, while longitudinal EHR data encode comorbidity, treatment history, and temporal dynamics. Existing multimodal approaches often select fusion strategies heuristically, and the interaction between fusion mechanism and temporal objective remains unexplored ~\cite{steinberg2024_motor_iclr,huo2024_tte_pretrain_med3d}. More recently, joint and attention-based fusion methods ~\cite{liang2024_mml_acmcs,SUN2023_JBI} have emerged as powerful strategies for integrating heterogeneous modalities. These approaches employ mechanisms such as cross-attention~\cite{Wang2023_TCBB}, co-attention~\cite{li2023_hmcat_tmi,ji2025_cbcat_jbhi}, and contrastive alignment~\cite{yang2022_csco_media,huang2024_maco_natcomm} to dynamically re-weight contributions from each modality based on contextual relevance.  However, foundation models for 3D imaging and clinical sequences are typically pretrained on generic objectives and are not optimized for TTE prediction, leaving latent representations unstructured for temporal risk modeling. Cross-modal alignment can reshape these embeddings toward temporally predictive signals, but existing approaches often rely on heuristic fusion strategies, and the interaction between alignment mechanism and survival objectives remains underexplored~\cite{steinberg2024_motor_iclr,huo2024_tte_pretrain_med3d}.

We propose a multimodal framework for time-to-event (TTE) prediction that systematically integrates supervised fusion strategies—contrastive alignment, cross-attention and bi-directional co-attention, and evaluate against traditional concatenation. Using mortality following pulmonary embolism (PE) and long-term cardiovascular outcomes as benchmark tasks, we demonstrate how the choice of fusion strategy impacts temporal risk modeling under distribution shift. These findings provide practical guidance for designing effective multimodal survival models that leverage complementary imaging and EHR information.

\section{Method}
\label{sec:method}
\begin{figure}[htb!]
  \centering
  \includegraphics[width=0.55\linewidth]{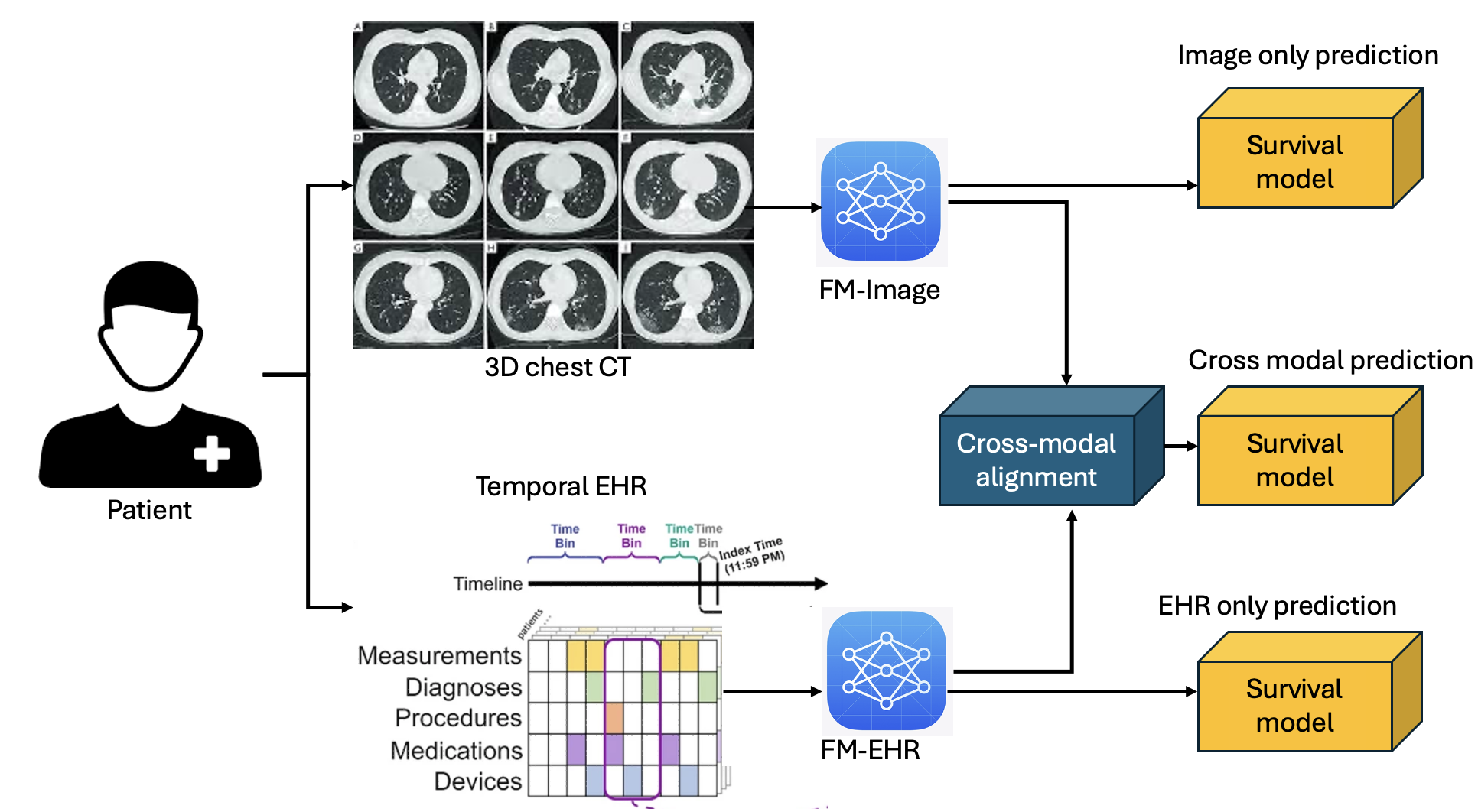}
  \begin{minipage}[b]{0.25\linewidth}
    \centering
    \includegraphics[width=\linewidth]{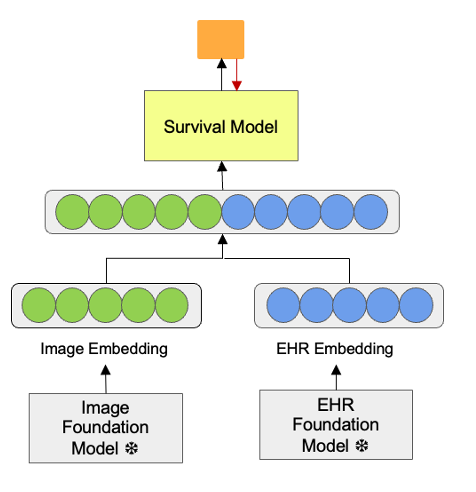}
    {\scriptsize (a) Traditional Concat}
  \end{minipage}
  \hfill
  \begin{minipage}[b]{0.25\linewidth}
    \centering
    \includegraphics[width=\linewidth]{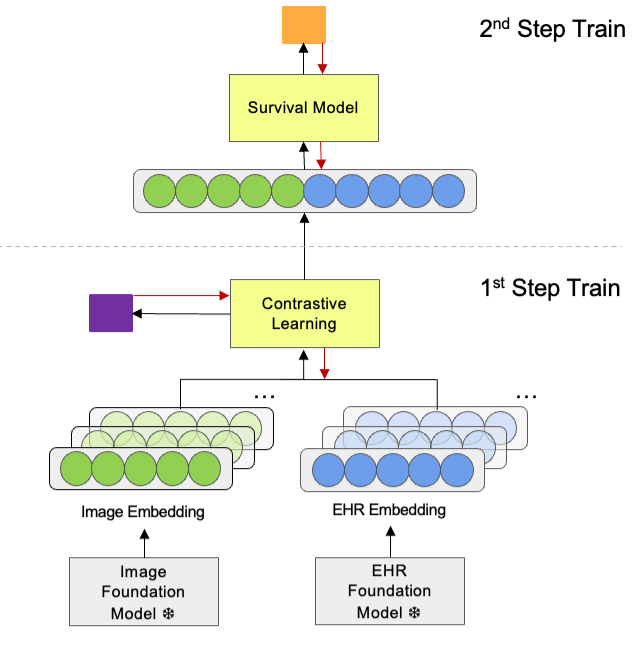}
    {\scriptsize (b) Proposed Contrastive}
  \end{minipage}
  \begin{minipage}[b]{0.25\linewidth}
    \centering
    \includegraphics[width=\linewidth]{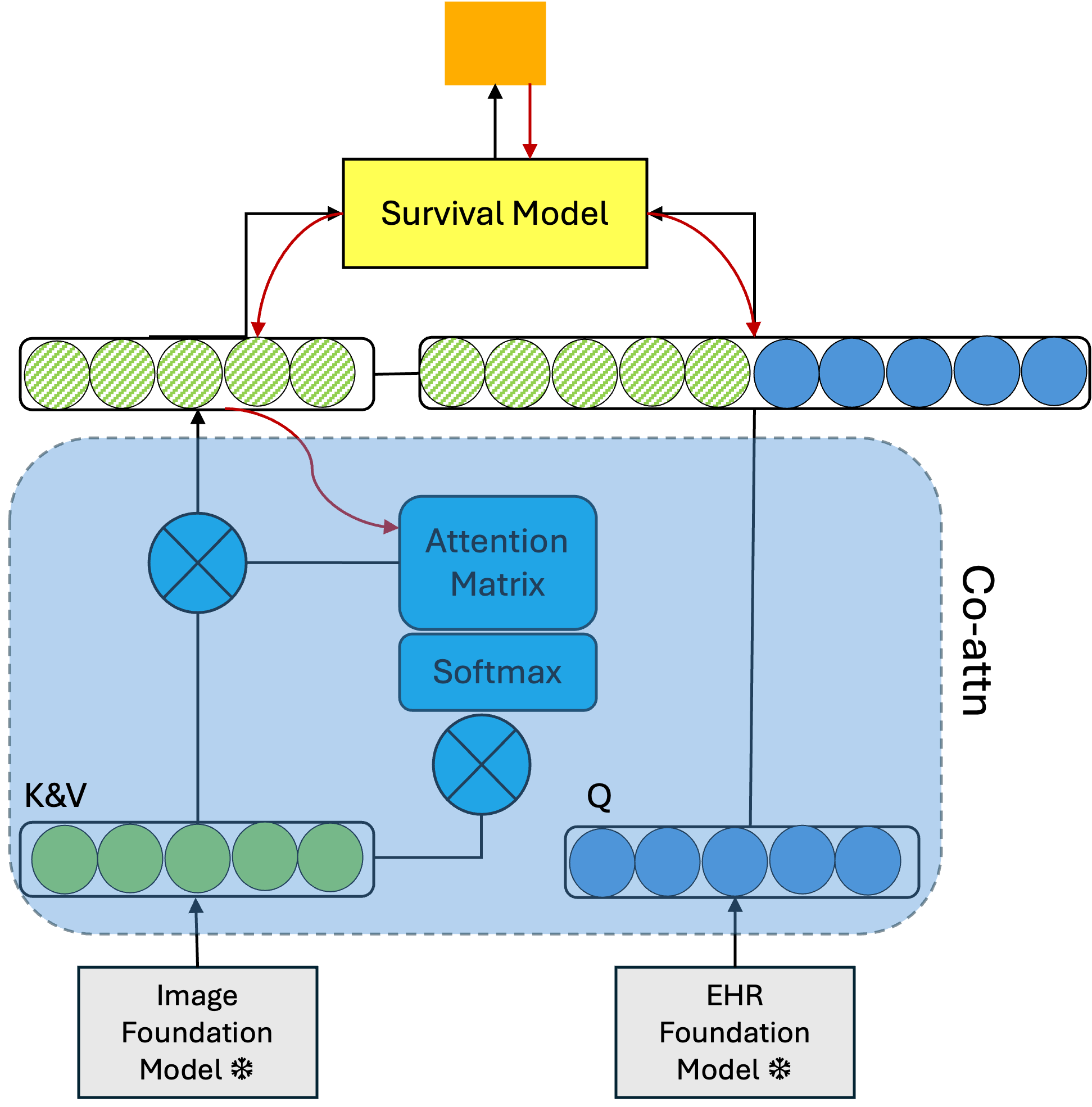}
    {\scriptsize (c) Proposed Co-Attention}
  \end{minipage}
  \begin{minipage}[b]{0.25\linewidth}\centering\includegraphics[width=\linewidth]{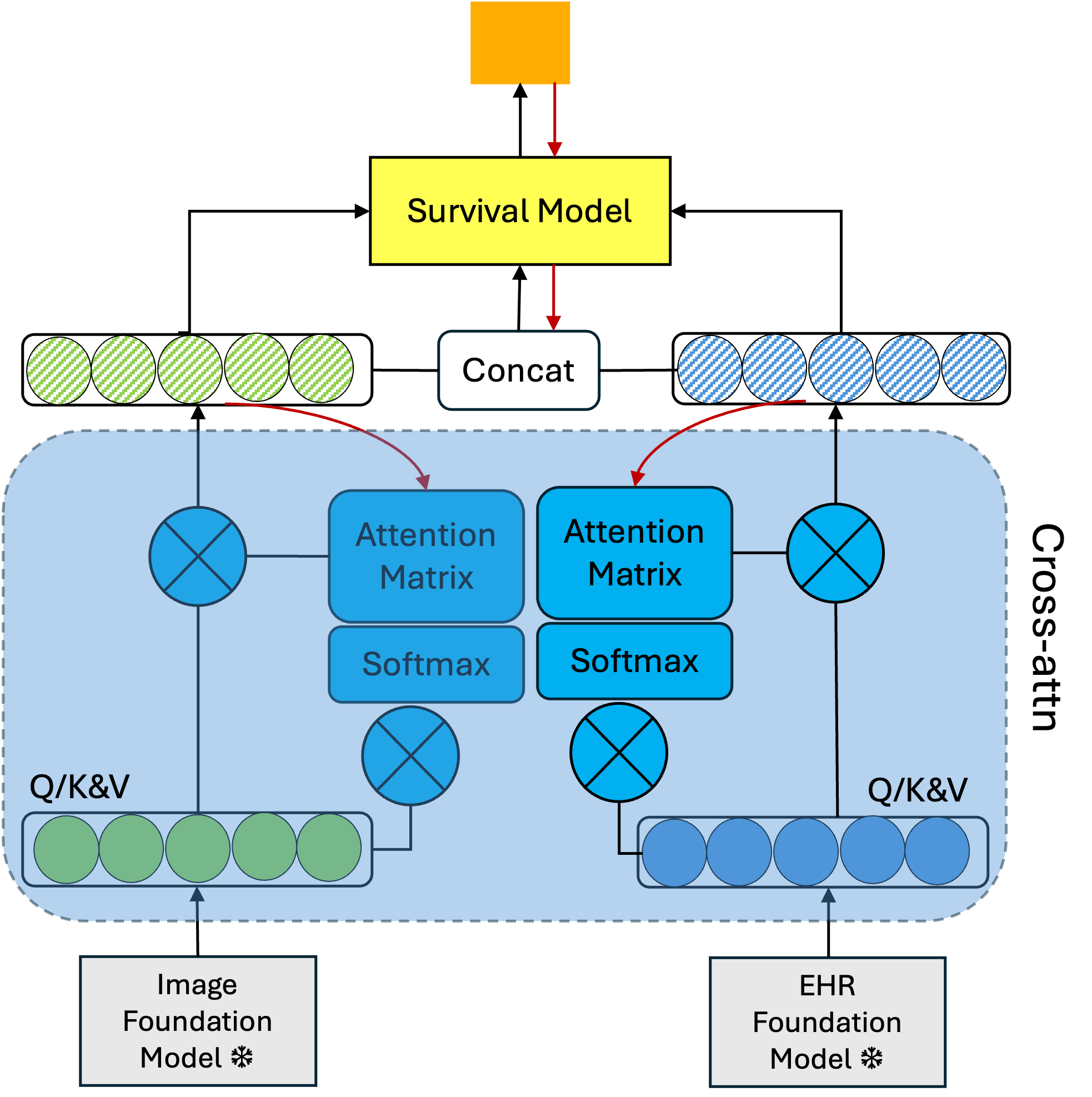}
    {\scriptsize (d) Proposed Cross-Attention}
  \end{minipage}
  
  \begin{minipage}[b]{0.65\linewidth}\centering\includegraphics[width=\linewidth]{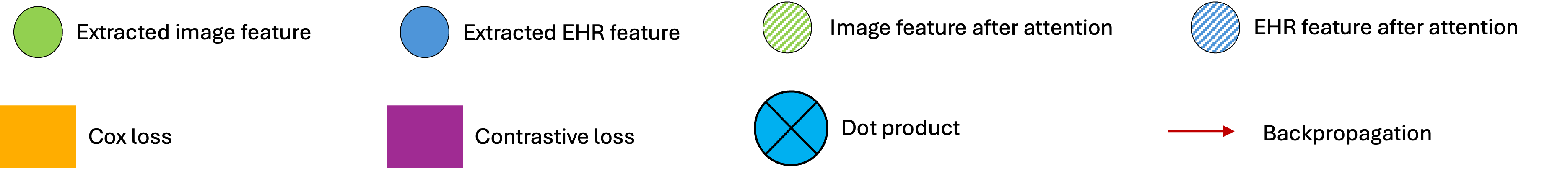}
   
  \end{minipage}
  \caption{ Overview of the multimodal survival framework and fusion strategies. Cross-modal alignment of chest CT and longitudinal EHR using domain-specific foundation encoders to produce shared embeddings for time-to-event prediction, with image-only and EHR-only baselines derived from the respective encoders. Fusion variants in latent space— (a) Traditional concatenation of embedding, Proposed - (b) contrastive learning, (c) co-attention, and (d) symmetric cross-attention.}
  \label{fig:models_combined}
\end{figure}

Figure~\ref{fig:models_combined} illustrates the multimodal fusion framework for time-to-event prediction. CT scans and EHR data are first encoded using domain-specific foundation models \cite{Neupane2025MedInsight}, producing image and clinical embeddings that are fused in latent space and optimized using a survival objective \cite{Monod2024TorchSurv}. Gradients are propagated only through the fusion and task-specific layers, while the pretrained foundation encoders remain frozen in all experiments.
\subsection{Foundation models - Image and EHR representation}
\emph{2D Medical Image Foundation Model – MedImageInsight:}
We used MedImageInsight (MII)~\cite{codella2024medimageinsightopensourceembeddingmodel}, a pretrained medical imaging foundation model that encodes 2D slices into $1\times1024$ embeddings.  For 3D CT volumes, axial slices are first selected with  cardiac structures.
Soft-tissue windowing is applied by clipping voxel intensities to $[-1350,150]$ HU for PE, and $[-125,225]$ HU for MACE. Each slice is independently encoded by MII, producing $N\times1024$ embeddings, which are averaged across the z-dimension to obtain a single $1\times1024$ volume-level representation. \emph{EHR Foundation Model – CLMBR Features:}
We use a pre-trained CLMBR-T-base model~\cite{guo2023_clmbr_srep}, an autoregressive Transformer trained via self-supervised next-code prediction on longitudinal structured EHR data. The model encodes time-ordered clinical codes and produces a 768-dimensional patient-level embedding. We utilize the publicly released Stanford-trained checkpoint and extract fixed embeddings without additional fine-tuning. As an alternative EHR representation, we also explore a manually curated 1-hot encoding, where task-specific features—including demographics, laboratory results, medications, diagnoses, and procedure codes—are selected and binarized into 1-hot vectors for downstream modeling.


\subsection{Survival Model: Time-to-Event (TTE) with cross modal alignment}
 For TTE modeling, we apply a multilayer perceptron (MLP) prediction head to either unimodal or fused embeddings and optimize a negative log-likelihood (NLL) survival objective that accounts for right-censored observations. Gradients are propagated through the fusion modules, enabling end-to-end optimization of the joint representation under the survival objective. Hidden-layer widths are scaled proportionally to the input embedding dimensionality, ensuring balanced model capacity across heterogeneous feature types. This formulation allows probabilistic estimation of event times by modeling both the hazard function and the censoring mechanism. Backpropagation of the NLL loss through the network, including the fusion modules, enables the shared representation to be shaped by temporally and clinically relevant features, improving predictive performance under both observed and censored conditions.
 
 \noindent \emph{Contrastive:} Inspired by CLIP style training~\cite{shah2022max_aaai,zhang2024learning_eccv}, we designed a 2-step training where we first align image and EHR embeddings using a symmetric contrastive objective. Given a batch of size $N$, the $i$-th image and $i$-th EHR form a positive pair, while all other combinations act as negatives. Similarity scores are defined as: $S_{ij}=\frac{\mathrm{sim}\!\left(h_i^{\mathrm{img}},h_j^{\mathrm{ehr}}\right)}{\tau},\quad
\mathcal{L} =-\frac{1}{N}\sum_{i=1}^{N}\left[
log\frac{e^{S_{ii}}}{\sum_{j=1}^{N}e^{S_{ij}}}
+\log\frac{e^{S_{ii}}}{\sum_{j=1}^{N}e^{S_{ji}}}
\right]$. After contrastive pre-alignment, the fused embeddings are used to train the time-to-event (TTE) survival model.

\noindent \emph{Cross-attention: } In an end-to-end learning, we fuse multimodal embeddings (imaging + EHR) using cross‑attention so that one modality can attend to important signals in the other, and then train this fused representation under a TTE objective. This lets the model simultaneously learn which features from each modality embedding are predictive of when an event is likely to occur in a \emph{supervised way}. Once we derived the foundation model encodings from image and ehr as: $h_{\text{img}} = f_{\text{enc,img}}(x_{\text{img}})$ and $h_{\text{ehr}} = f_{\text{enc,ehr}}(x_{\text{ehr}})$, we calculated as fused feature $h_{\text{fused}}$ with cross attention as - $h_{\mathrm{fused}} = \mathrm{concat}\!\bigl(h_{\mathrm{ehr}},\,A_{\mathrm{ehr}\leftarrow\mathrm{img}},\,h_{\mathrm{img}},\,A_{\mathrm{img}\leftarrow\mathrm{ehr}}\bigr)$, where 
\begin{equation*}
Q_{\mathrm{ehr}} = W_Q^{(e)}\,h_{\mathrm{ehr}}, \,
K_{\mathrm{img}} = W_K^{(e)}\,h_{\mathrm{img}}, \,
V_{\mathrm{img}} = W_V^{(e)}\,h_{\mathrm{img}} 
\end{equation*}
\begin{equation*}
A_{\mathrm{ehr}\leftarrow\mathrm{img}} = \mathrm{softmax}\!\Bigl(\frac{Q_{\mathrm{ehr}}\,K_{\mathrm{img}}^\top}{\sqrt{d_k}}\Bigr)\;V_{\mathrm{img}}
\end{equation*}
\begin{equation*}
Q_{\mathrm{img}} = W_Q^{(i)}\,h_{\mathrm{img}}, \,
K_{\mathrm{ehr}} = W_K^{(i)}\,h_{\mathrm{ehr}}, \,
V_{\mathrm{ehr}} = W_V^{(i)}\,h_{\mathrm{ehr}} 
\end{equation*}
\begin{equation*}
A_{\mathrm{img}\leftarrow\mathrm{ehr}} = \mathrm{softmax}\!\Bigl(\frac{Q_{\mathrm{img}}\,K_{\mathrm{ehr}}^\top}{\sqrt{d_k}}\Bigr)\;V_{\mathrm{ehr}} 
\end{equation*}
and \( W_Q^{(e)}, W_K^{(e)}, W_V^{(e)} \) are projection matrices for the EHR-to-Image cross-attention mechanism, where queries are derived from EHR embeddings, and keys and values are from image embeddings. Specifically, \( Q_{\mathrm{ehr}} = W_Q^{(e)} h_{\mathrm{ehr}} \), \( K_{\mathrm{img}} = W_K^{(e)} h_{\mathrm{img}} \), and \( V_{\mathrm{img}} = W_V^{(e)} h_{\mathrm{img}} \). The reverse, Image-to-EHR cross-attention, follows the same notation but with roles of modalities swapped. The outputs of these attention mechanisms, \( A_{\mathrm{ehr} \leftarrow \mathrm{img}} \) and \( A_{\mathrm{img} \leftarrow \mathrm{ehr}} \), are concatenated with their respective modality embeddings to form the fused feature vector \( h_{\text{fused}} \). This fused representation is then utilized for time-to-event (TTE) prediction, formulated as:
$\mathcal{L_{TTE}} = - \sum_{i=1}^N \log p\left( T_i, \delta_i \mid h_{\text{fused},i} \right) + \lambda \|\theta\|_2^2$, where \( T_i \) denotes the observed or censored time for subject \( i \), \( \delta_i \) is the event indicator, \( \lambda \) is the regularization hyperparameter, and \( \theta \) represents the model's trainable parameters across projection layers and attention mechanisms.

\noindent \emph{Co-attention: } 
To preserve the primary knowledge from one modality and use other as a guiding modality, we adapted the co-attention mechanism. In this model, one modality (e.g. \(h_{\mathrm{ehr}}\) ) serves as the query, while the other \(h_{\mathrm{img}}\) functions as both the key and value in the co‑attention computation:

\[
\text{CoAttn}_{\mathrm{ehr} \to \mathrm{img}}(h_{\mathrm{ehr}},\, h_{\mathrm{img}}) = \mathrm{softmax}\!\left( \frac{Q_{\mathrm{ehr}}\, K_{\mathrm{img}}^\top}{\sqrt{d_k}} \right)\; V_{\mathrm{img}}
\]

\noindent
where $Q_{\mathrm{ehr}} = W_Q^{(e)}\,h_{\mathrm{ehr}}$, $K_{\mathrm{img}} = W_K^{(e)}\,h_{\mathrm{img}}$, $V_{\mathrm{img}} = W_V^{(e)}\,h_{\mathrm{img}}$. The attention scores are computed by comparing the projected query \( Q = W_q^{\text{ehr}}\,h_{\text{ehr}} \) with the projected key \( K = W_k^{\text{img}}\,h_{\text{img}} \), scaled by \( \sqrt{d_k} \) for numerical stability. After that, we apply the softmax operation to obtain a co‑attention matrix \( A_{\mathrm{coattn}} = \mathrm{softmax}\left( \frac{Q\,K^\top}{\sqrt{d_k}} \right) \), which assigns importance weights over different regions of image modality. These weights are then used to compute a weighted sum of the image features, producing a refined representation \( \hat{h}_{\text{img}} \) as: $\hat{h}_{\text{img}} = A_{\mathrm{coattn}}\, (W_v^{\text{img}}\, h_{\text{img}})
\quad = \quad \mathrm{softmax}\left( \frac{W_q^{\text{ehr}}\, h_{\text{ehr}}\, (W_k^{\text{img}}\, h_{\text{img}})^\top}{\sqrt{d_k}} \right) \, W_v^{\text{img}}\, h_{\text{img}}
$, where \( W_q^{\text{ehr}}, W_k^{\text{img}}, W_v^{\text{img}} \in \mathbb{R}^{d_k \times d_k} \) are trainable weight matrices. The value term is derived from image as \( V = W_v^{\text{img}}\, h_{\text{img}} \), which ensures that the model focuses on the most informative regions of the imaging modality given the EHR features. However, the image and EHR modalities can ideally swapped since the attention is being learn in embedding space. Similar to cross attention, a fused embedding is generated after the incorporating the coattention ($h_{\mathrm{fused}}  = concat(h_{img}, A_{\mathrm{coattn}})$)  which is finally fed into the TTE to optimize the overall loss: $L_{TTE}$.


\section{Results}
\noindent \emph{Clinical use-case 1: Major adverse cardiac event (MACE) prediction} — Traditional MACE risk models rely on predefined tabular variables (e.g., PCE) or imaging alone (e.g., CAC) to predict risk of myocardial infarction, ischemic stroke, heart failure and cardiac mortality, limiting personalization and underutilizing longitudinal EHR data. While chest CT captures detailed anatomy, EHR provides temporal clinical context, though often incomplete in asymptomatic patients. Integrating both modalities via foundation models enables complementary, personalized risk stratification. We developed the model using retrospective multi-site data (3,947 patients; 4,133 thoracic CTs; 46\% MACE rate) and evaluated generalization on ED patients undergoing routine non cardiac chest CTs(665 patients; 682 CTs; 28\% MACE rate). Temporal EHR features were extracted per AHA guidelines, including demographics, vitals, labs, comorbidities, and medications.
\noindent \emph{Clinical use-case 2: Mortality prediction in pulmonary embolism (PE)} — PE carries substantial mortality risk (1–35\% depending on severity), underscoring the need for accurate stratification. Existing tools (e.g., PESI) emphasize clinical variables but rarely integrate high dimensional imaging data. We assembled a private dataset of all positive acute PE diagnosis confirmed CTA pulmonary angiographic studies (3,764 patients; 4,542 studies; 7\% mortality for 1 year) with linked temporal EHR data mapped to OMOP CDM for standardized learning. External validation was performed on randomly sampled all positive acute PE INSPECT dataset~\cite{huang2023inspect} (396 patients; 435 studies; 19\% mortality for 1 year), enabling robust multimodal mortality prediction across sites.

\noindent \emph{Quantitative Performance} - As baseline multimodal integration strategies, we evaluated simple feature concatenation (Fig.~\ref{fig:models_combined}) where concatenation directly combines independently pre-trained image and EHR embeddings into a joint representation. This establish baseline multimodal integration strategies against which 2-step contrastive and attention-based fusion methods are compared. The frozen image-only encoder with supervised training of survival layers is used as the unimodal baseline, given the heterogeneity in EHR encoding strategies (e.g., one-hot representations versus CLMBR-based embeddings), which could otherwise introduce confounding variability in baseline comparisons.
\begin{table}[!htb]
\centering
\caption{\footnotesize Prediction performance for PE and MACE reported as mean (95\% CI) for Internal (Institution A) and External (Inspect) cohorts. Unimodal baselines (Image-only and EHR-only) are shown separately from multimodal fusion models. $^{*}$ indicates $p<0.05$ versus Image-only, and $^{+}$ indicates $p<0.05$ versus linear concatenation fusion as ref. baseline. Best performance within each cohort and outcome is shown in \textbf{bold}.}
\label{tab:pe_image_stat}
\setlength{\tabcolsep}{4pt}
\renewcommand{\arraystretch}{1.15}
\footnotesize
\resizebox{\textwidth}{!}
{%
\begin{tabular}
{lcc|cc}
\toprule
& \multicolumn{2}{c|}{\textbf{Internal}} 
& \multicolumn{2}{c}{\textbf{External}} \\
\cmidrule(lr){2-3} \cmidrule(lr){4-5}
\textbf{Model} & \textbf{PE} & \textbf{MACE} 
& \textbf{PE} & \textbf{MACE} \\
\midrule

\multicolumn{5}{l}{\textbf{Image-Only Baseline}} \\
Image only\textit{(unimodal ref.)} & 0.837 (0.828--0.847) & 0.742 (0.737--0.748) & 0.710 (0.697--0.722) & 0.725 (0.718--0.732) \\

\midrule
\multicolumn{5}{l}{\textbf{EHR-Only (Unimodal Structured Data)}} \\
\addlinespace[0.2em]
\textit{One-hot Representation} \\
EHR only & 0.748$^{*}$ (0.735--0.760) & 0.732$^{*}$ (0.725--0.739) & 0.620$^{*}$ (0.610--0.630) & 0.691$^{*}$ (0.685--0.698) \\

\addlinespace[0.2em]
\textit{CLMBR Representation} \\
EHR only & 0.775$^{*}$ (0.763--0.787) & 0.502$^{*}$ (0.498--0.505) & 0.641$^{*}$ (0.628--0.654) & 0.497$^{*}$ (0.497--0.498) \\

\midrule
\multicolumn{5}{l}{\textbf{Multimodal Fusion Models}} \\
\addlinespace[0.2em]
\textit{One-hot Representation} \\
Concatenation \textit{(cross-modal ref.)} & 0.819$^{*}$ (0.810--0.828) & 0.790$^{*}$ (0.786--0.794) & 0.717 (0.708--0.726) & 0.733 (0.725--0.740) \\
Contrastive Learning & 0.847$^{+}$ (0.840--0.854) & 0.794 (0.790--0.798) & 0.738$^{*+}$ (0.724--0.752) & 0.722 (0.714--0.729) \\
Cross-attention & 0.846$^{+}$ (0.839--0.854) & \textbf{0.796}$^{*}$ (0.791--0.801) & 0.719 (0.709--0.729) & 0.731 (0.724--0.738) \\
Co-attention (image guide) & 0.839$^{+}$ (0.830--0.848) & 0.791 (0.786--0.796) & 0.693$^{*+}$ (0.685--0.700) & \textbf{0.740}$^{*}$ (0.735--0.746) \\
Co-attention (EHR guide) & 0.784$^{*+}$ (0.774--0.795) & 0.791 (0.787--0.795) & 0.653$^{*+}$ (0.641--0.664) & 0.736$^{*}$ (0.730--0.741) \\

\addlinespace[0.3em]
\textit{CLMBR Representation} \\
Concatenation \textit{(cross-modal ref.)} & 0.846 (0.837--0.855) & 0.757$^{*}$ (0.751--0.763) & 0.734 (0.724--0.744) & 0.721 (0.712--0.730) \\
Contrastive Learning & \textbf{0.862}$^{*+}$ (0.856--0.868) & 0.759$^{*}$ (0.754--0.764) & \textbf{0.743}$^{*+}$ (0.731--0.754) & 0.727 (0.721--0.734) \\
Cross-attention & 0.846 (0.838--0.854) & 0.762$^{*}$ (0.757--0.767) & 0.723 (0.707--0.739) & 0.725 (0.718--0.731) \\
Co-attention (image guide) & 0.842 (0.836--0.848) & 0.741 (0.733--0.749) & 0.712 (0.701--0.723) & 0.724 (0.715--0.732) \\
Co-attention (EHR guide) & 0.824$^{+}$ (0.814--0.834) & 0.754 (0.749--0.759) & 0.724$^{*}$ (0.715--0.733) & 0.726 (0.718--0.734) \\

\bottomrule
\end{tabular}
}
\end{table}

\begin{figure}[!htb]
  \centering
  \includegraphics[width=0.7\linewidth]{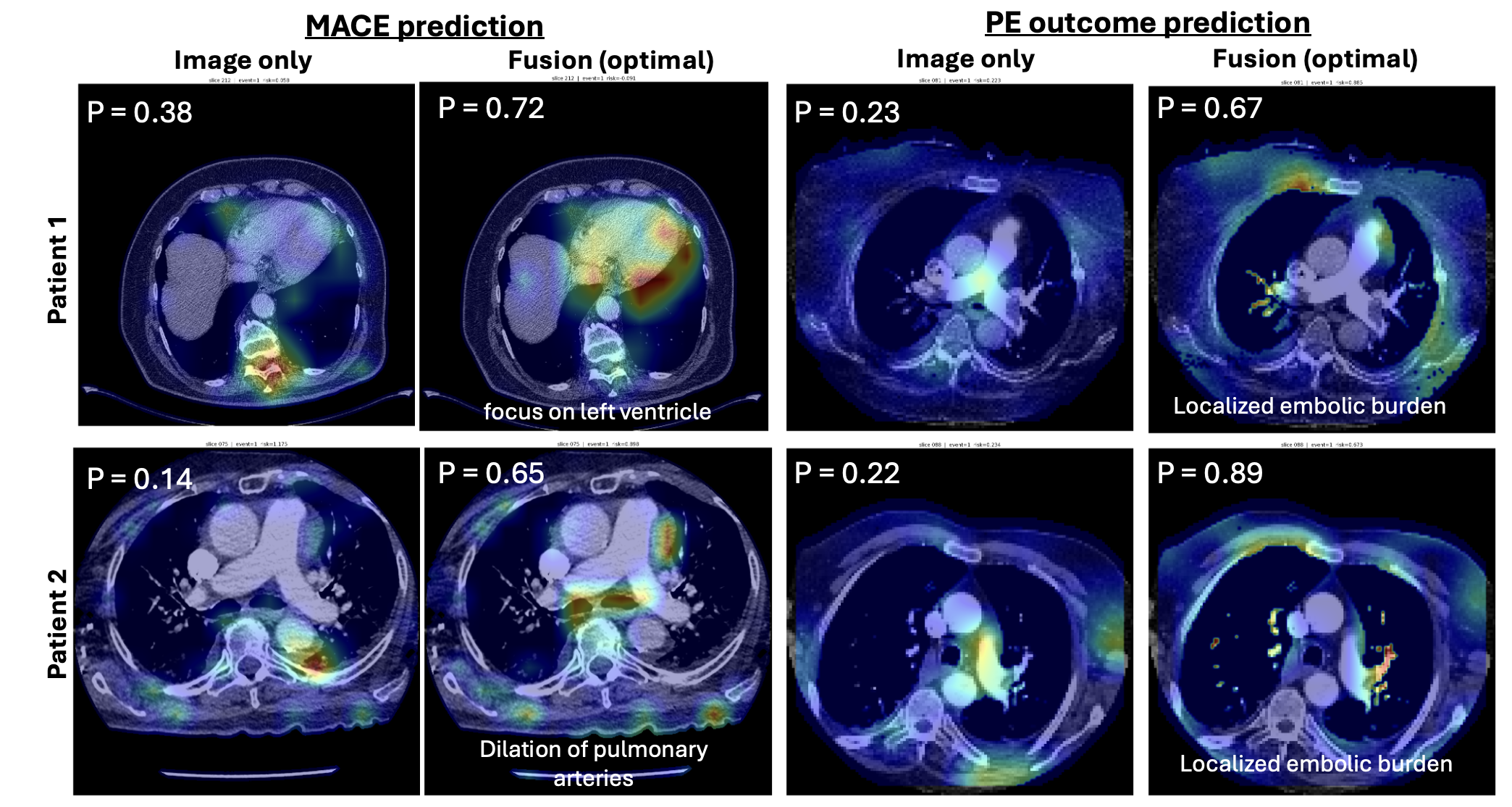}
  \caption{Comparative saliency maps for MACE and pulmonary embolism (PE) outcome prediction using image-only and multimodal fusion models. \emph{P} denotes the predicted risk score for patients who subsequently developed adverse events.}
  \label{fig:saliency_map}
\end{figure}

Across both PE and MACE tasks, multimodal fusion models generally improved prediction performance over unimodal baselines, with statistically significant gains observed across several settings. EHR-only models performed significantly worse than the image-only reference for both PE and MACE (p<0.05). For PE prediction, 2-step contrastive learning consistently achieved the strongest results, with the CLMBR-based contrastive model demonstrating the best overall performance internally (AUC 0.862) and externally (AUC 0.743), significantly outperforming both image-only and linear concatenation (p<0.05). For MACE, improvements were more modest, though cross-attention (one-hot) achieved the highest internal performance (AUC 0.796, p<0.05 vs image-only) and image-guided co-attention achieved the best external performance (AUC 0.740, p<0.05 vs image-only). One-hot representations provided more stable MACE prediction, likely due to the use of hand-curated PREVENT features employed in current clinical risk scoring~\cite{thomsen2001new}. Overall, contrastive multimodal fusion—particularly with CLMBR representations—provided the most consistent and statistically robust improvements, especially for PE and external generalization. Statistically significant gains over traditional linear concatenation of the foundational feature space underscore the value of task-driven cross-modal alignment, and performance trends suggest that optimal fusion strategies are task-dependent and performs better when trained with supervised objective.
We analyzed gradient-based saliency maps to assess whether multimodal alignment altered spatial attention. Compared to image-only models, cross-attention fusion shifted attention toward clinically relevant regions (Fig.~\ref{fig:saliency_map}). For PE mortality prediction, the image-only model primarily highlighted structure of pulmonary arteries, whereas the fusion model localized embolic burden. For MACE risk estimation, the fusion model emphasized pulmonary artery dilation in addition to cardiac structures, yielding more confident predictions. Although saliency maps do not establish causality, these findings suggest that integrating EHR information guides imaging features toward task-relevant anatomy.

\section{Discussion}
Vision–language pretraining aligns modalities using self-supervised objectives (e.g., contrastive losses), but such alignment does not necessarily transfer optimally to complex time-to-event (TTE) prediction, where prognostic signals may differ from pretraining objectives. While concatenation of pre-aligned spaces provide useful baseline, they may fail to capture fine-grained task-relevant interactions. In contrast, supervised attention-based fusion enables frozen foundation embeddings to be selectively aligned for survival prediction without full fine-tuning.

Across MACE and PE mortality tasks, both modality importance and optimal fusion strategy were task-dependent. CT imaging was more discriminative for MACE, reflecting the prognostic value of cardiac morphology and coronary calcification, whereas longitudinal EHR features were more informative for PE mortality, which depends on systemic comorbidities and treatment context. EHR representation choice further influenced performance: curated one-hot cardiovascular variables provided stable MACE prediction, while CLMBR foundational embeddings, which capture broader comorbidity patterns, better predicted PE-related mortality but underperformed for MACE. Results underscore a limitation of foundation models—representations optimized for general sequence modeling may fail to retain task-specific survival signals without targeted alignment.

Attention-based strategies surpassed simple concatenation by adaptively weighting modalities, enhancing discrimination and robustness under modality imbalance. Bidirectional cross-attention achieved the highest internal performance, while co-attention improved external generalization when modality importance was asymmetric, highlighting that optimal fusion is task- and cohort-dependent. CLMBR-based 2-step contrastive fusion provided the most consistent and statistically robust gains, particularly for PE and external validation. Saliency analyses supported these findings: fusion shifted attention toward clinically relevant regions, localizing embolic burden for PE and emphasizing pulmonary artery dilation and cardiac morphology for MACE, demonstrating that EHR integration guides imaging representations to task-relevant anatomy.

External validation revealed persistent generalization gaps, likely driven by heterogeneity in imaging protocols, population characteristics, and EHR completeness, underscoring the need for domain adaptation and harmonization strategies for cross-site deployment with foundational backbone. Additional limitations include cohort size, event-rate imbalance, potential residual confounding in retrospective data. Prospective validation and calibration assessment will be essential before clinical translation. Experimental results indicate that even with powerful self-supervised foundation models, supervised, task-aware cross-modal learning is critical for robust and clinically meaningful survival modeling in multimodal settings.
\bibliographystyle{splncs04}
\bibliography{main}
\end{document}